\documentclass{article}

    \PassOptionsToPackage{numbers, compress}{natbib}

\usepackage[final]{neurips_2022}

\usepackage[utf8]{inputenc} %
\usepackage[T1]{fontenc}    %
\usepackage{hyperref}       %
\usepackage{url}            %
\usepackage{booktabs}       %
\usepackage{amsmath}        %
\usepackage{amsfonts}       %
\usepackage{microtype}      %
\usepackage{xcolor}         %
\usepackage{graphicx}
\usepackage[capitalise,nameinlink]{cleveref}
\usepackage{paralist}
\usepackage{caption}        %
\usepackage{subcaption}     %
\usepackage{scalerel}       %
\usepackage{tikz}           %
\usepackage{bm}             %
\usepackage[export]{adjustbox} %
\usepackage{multirow}       %
\usepackage{tabularx}       %
\usepackage{wrapfig}
\usepackage{pifont}
\usepackage{physics}        %
\usepackage{rotating}
\usepackage{algorithmic}
\usepackage{algorithm}

\newcommand{\Shift}{\texttt{Shift}}
\newcommand{\LArc}{\texttt{LArc}}
\newcommand{\RArc}{\texttt{RArc}}
\newcommand{\Reduce}{\texttt{Reduce}}
\title{Transition-based Abstract Meaning Representation Parsing with Contextual Embeddings}
\newcommand{\amrberger}{\textsc{AmrBerger}}
\newcommand{\amreager}{\textsc{AmrEager}}

\author{%
  Yichao Liang\thanks{Written in 2020 during an internship at Shay Cohen's lab, The University of Edinburgh.}\\
  Department of Computer Science\\
  University of Oxford\\
  \texttt{ycliang6@gmail.com} \\
}

\begin{document}

\maketitle
\begin{abstract}
The ability to understand and generate languages sets human cognition apart from other known life forms'. We study a way of combing two of the most successful routes to meaning of language--statistical language models and symbolic semantics formalisms--in the task of semantic parsing. 
Building on a transition-based, Abstract Meaning Representation (AMR) parser, \textsc{AmrEager}, we explore the utility of incorporating pretrained context-aware word embeddings--such as BERT\cite{devlin2018bert} and RoBERTa \cite{liu2019roberta}--in the problem of AMR parsing, contributing a new parser we dub as \textsc{AmrBerger}.
Experiments find these rich lexical features alone are not particularly helpful in improving the parser's overall performance as measured by the SMATCH score when compared to the non-contextual counterpart, while additional concept information empowers the system to outperform the baselines. 
Through lesion study, we found the use of contextual embeddings helps to make the system more robust against the removal of explicit syntactical features.
These findings expose the strength and weakness of the contextual embeddings and the language models in the current form, and motivate deeper understanding thereof.
\end{abstract}

\section{Introduction}
Semantic formalism attempts to, from one perspective represent the meaning of natural language while disregarding its superficial features, and from another precisely represent our knowledge of the state of affairs in the world.
Abstract Meaning Representation (AMR) \citep{banarescu2013abstract} is a sentence-level semantic formalism of languages that attempts to do a practical, replicable amount of canonicalization of sentences, while captures many aspects of meaning in a simple data structure--directed graphs.
It encodes information about semantic relations, named entities, co-reference, negation, and modality, which are usually independently addressed in natural language processing (NLP).
This representation has been shown to be useful in many downstream NLP tasks, including text summarization \cite{liu2015smith, dohare2017text}, question answering \cite{mitra2016addressing}, machine translation \cite{jones2012semantics, song2019semantic}, and more.

To use AMR in other NLP tasks, parsers are needed to generate AMR graphs automatically from linguistic sentences. 
In recent years, many parsers have been developed \cite{naseem2019rewarding, zhou2020amr, konstas2017neural, lyu2018amr}. Among them, graph-based and transition-based are the two main families; 
graph-based parsing decomposes the task into concept and relation identification, then determines a full AMR graph with decoding algorithms, e.g., maximum spanning tree, 
while transition-based systems build up AMR graphs incrementally by applying a sequence of transition actions. 
The parser of our study, \textsc{AmrEager}, is an instance of the greedy, transition-based approaches (greedy in the sense that the parser provides a single choice of action at each state, and a single parse is returned in the end, without any exploration and backtracking).

To predict the transition action at each time step, a parser takes into account lexical information--especially the semantics of words--along with possibly syntactical features from the parts of the sentence it will parse and/or has parsed. 
To supply this, word embeddings stemmed from the distributional hypothesis \cite{harris1954distributional} are employed. 
For example, GloVe \cite{pennington2014glove} is a widely-used model for obtaining context-independent word features by computing global word-word co-occurrence statistics of a  large corpus, 
BERT \cite{devlin2018bert} and RoBERTa \cite{liu2019roberta} embeddings are a more modern, contextual approach obtained from bidirectional transformer models \cite{vaswani2017attention}, which have demonstrated strong signaling ability on a variety of NLP tasks.

\textsc{AmrEager} was originally developed to use context-independent word embeddings, e.g. GloVe, and performed poorly at, in particular, the Semantics Role Labelling (SRL) metric, a crucial subtask of AMR parsing (as discussed in \citet{damonte-17}). 
On the other hand, \citet{shi2019simple} show a simple BERT model can achieve state-of-the-art in the SRL task, motivating us to experiment with equipping contextual embedding (BERT, RoBERTa) for \textsc{AmrEager}, 
which is the main contribution of this paper.

The paper is organized as follows. \cref{sec:background} discusses the background of AMR, \textsc{AmrEager} and contextual embeddings. \cref{sec:design} describes the design of our approach, and \cref{sec:experiments} details the experimental design and results.

\section{Background}\label{sec:background}
\begin{figure}[ht]
    \centering
    \begin{subfigure}[b]{.49\textwidth}
        \includegraphics[width=.8\textwidth]{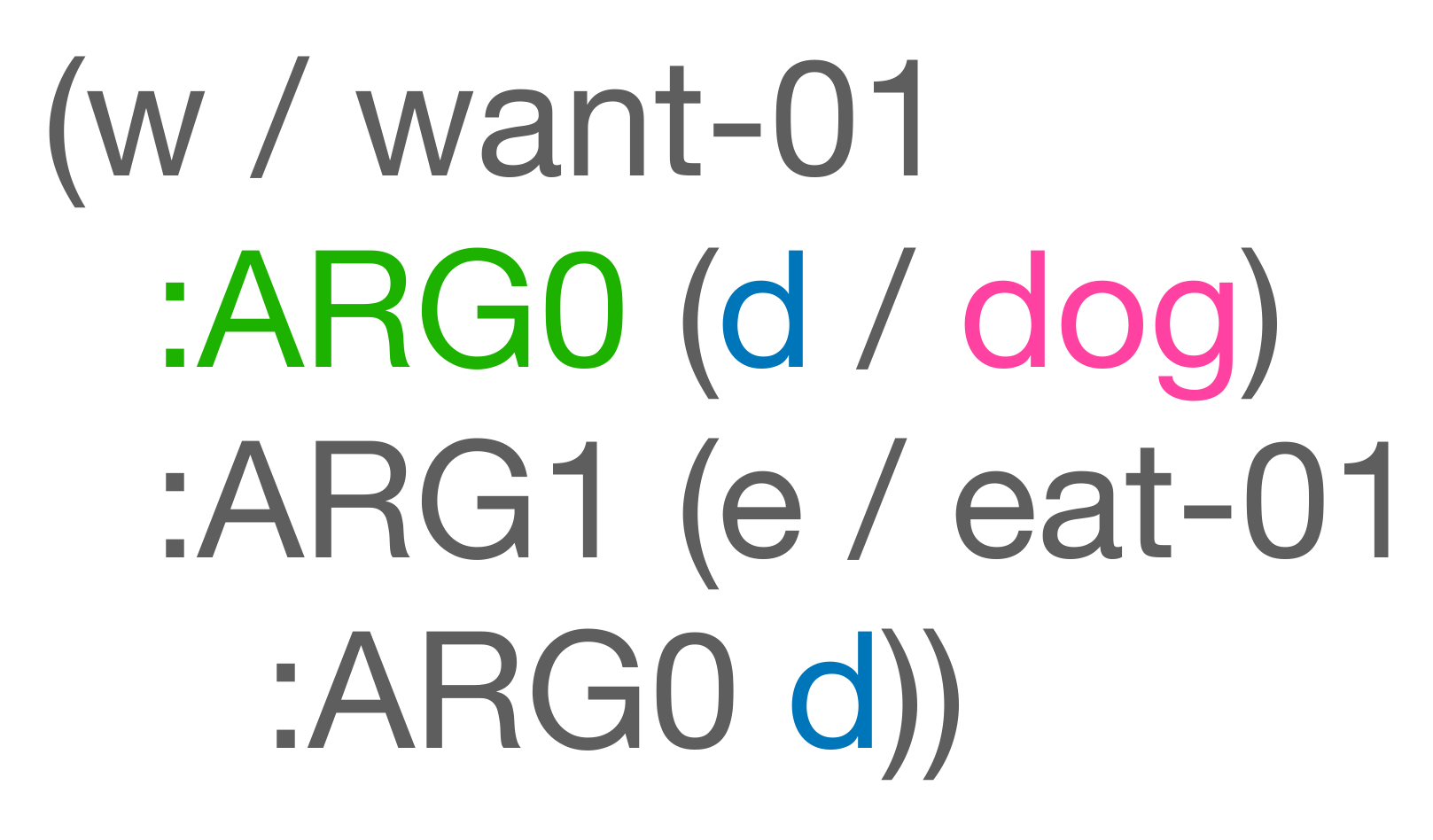}
        \caption{}
    \end{subfigure}
    \begin{subfigure}[b]{.49\textwidth}
        \includegraphics[width=.8\textwidth]{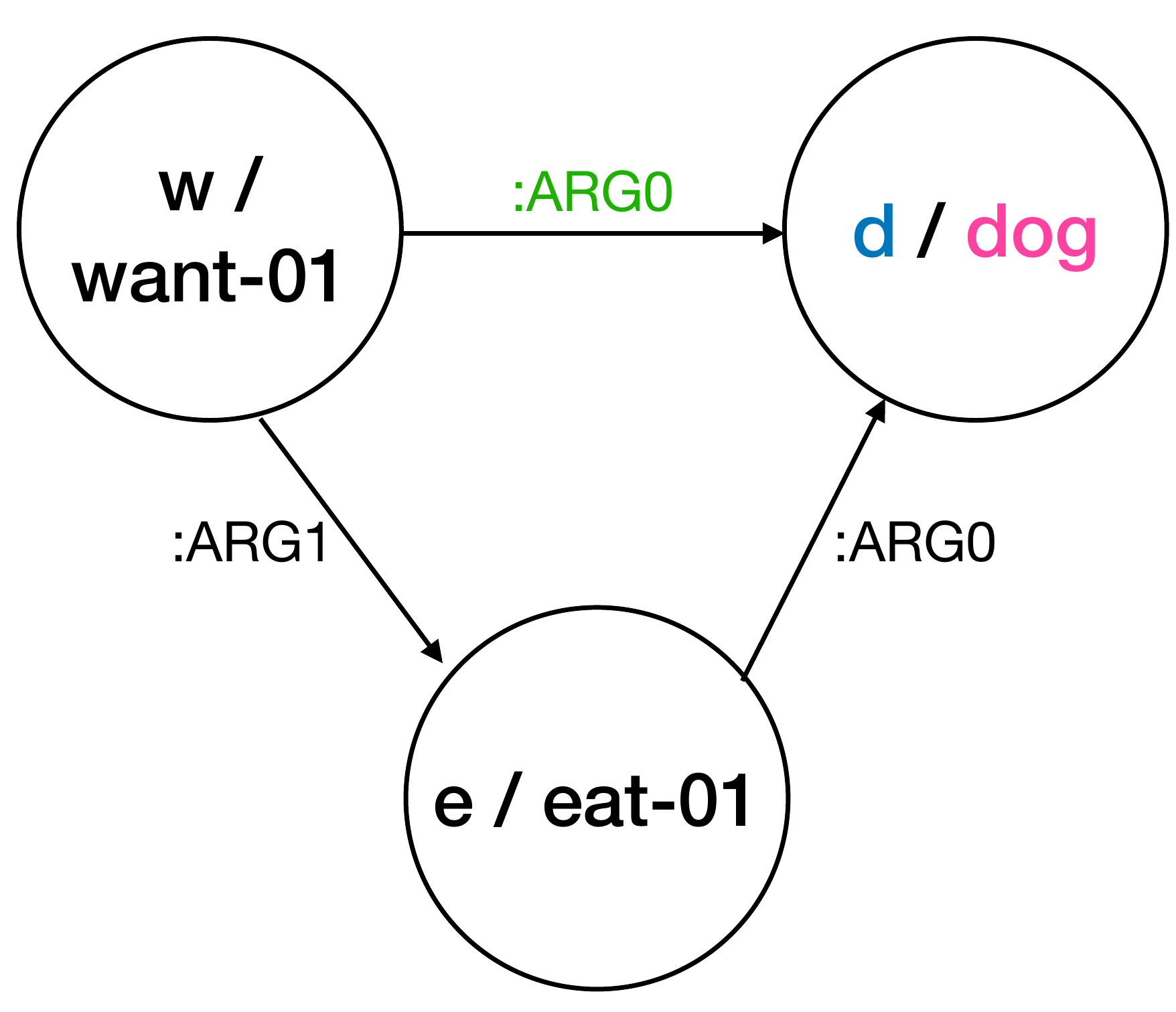}
        \caption{}
    \end{subfigure}
    \caption{(a) shows an AMR of sentence ``The dog wants to eat" where ``d / dog" means ``d is an instance of the concept dog" and ``ARG-0" represents a relation. (b) shows the correspondingg graph representation of the AMR structure in (a).}
    \label{fig:amr_exp}
\end{figure}

\begin{figure}[ht]
    \centering
    \fbox{\includegraphics[width=.5\textwidth]{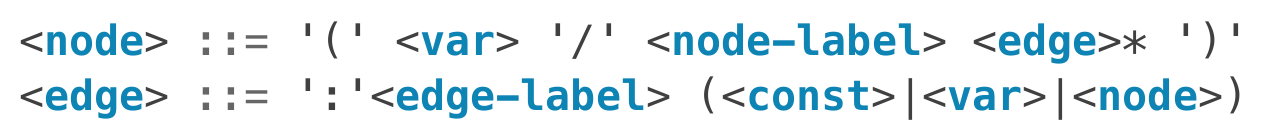}}
    \caption{The rough grammar of AMR}
    \label{fig:amrgrammar}
\end{figure}

\paragraph{AMR} In the AMR framework, each sentence is annotated with the PENMAN notation \cite{matthiessen1991text}, a serialized representation for directed, rooted graphs, similar to Lisp S-Expressions. 
As an example, \cref{fig:amr_exp} shows an AMR of the sentence ``The dog wants to eat" in PENMAN notation and its corresponding graph, respectively. 
In the graph, each node is labeled by a variable and a concept (e.g., ``d / dog") or a constant (e.g. negation, a quantity), while each edge represents a relation, similar to PropBank arguments \cite{kingsbury2002treebank}. 
The representation can be roughly expressed as a grammar in Backus–Naur form as in \cref{fig:amrgrammar}. 
Notably, there isn't always a straightforward alignment between words in a sentence and its AMR notes; the mapping from a graph node to a word is neither injective nor surjective (i.e., each word can correspond to zero or more nodes). 
However, alignments are often still required in training AMR parsers, including for \textsc{AmrEager}. 
JAMR \cite{flanigan2014discriminative} is a popular aligner based on a set of heuristics rules and a greedy search process. And it's used in this work to generate alignments.

\paragraph{Notation}
Let $[n]$ represent the set $\{1,...,n\}\in\mathbb{Z}$. Following the notation of \cite{damonte-17}, each AMR structure is defined as a 3-tuple $(G,x,\pi)$, 
where $x = (x_1,\dots,x_n)$ is a sentence with $n$ tokens.

\paragraph{The \textsc{AmrEager} Transition System}

Taking inspiration from the dependency parsing transition system \textsc{ArcEager} \cite{nivre2004incrementality}, and insight that AMR parsing differs from English dependency parsing mainly in the following three aspects:
\begin{inparaenum}[(i)]
\item AMR parsing requires the mapping from word tokens to AMR nodes; 
\item the existence non-projectivity in the graph (in 51\% of the sampled sentences); 
\item the existence of reentrant nodes (in 93\% of the sampled sentences \cite{damonte-17}). 
\end{inparaenum}

\textsc{AmrEager} is proposed by \citet{damonte-17} as a data-driven \emph{transition system} for generating ARM graphs as sequences of transition actions, with sentence tokens and their dependency graphs as input.
A run of the transition system can be completely characterized by a sequence of \emph{configurations} (the state of the parser, the stack, the input buffer of word tokens, and the set of relations representing an AMR graph) and the transition actions among them.

\textsc{AmrEager} makes the following three assumptions when perform parsing:
\begin{inparaenum}[(i)]
\item AMR graphs are directed acyclic graphs (DAGs). (In practice, AMR graphs with cycles are rare, but do exist.); 
\item Each node in a graph aligns to at most one sentence token, meaning that the node represents a part of the concept expressed by the token. Consequently, co-references of nodes from tokens are not considered; 
\item Co-references between nodes are ignored because reentrancy happens mostly only between siblings of a control verb. 
\end{inparaenum}

The transition system consists of four components: a \emph{parser} (consisting of an oracle used during training for generating training/validation dataset for all components of the system), a \emph{buffer} for storing input, a \emph{stack} for storing partially completed graph, and a \emph{list} for recording parsing actions.

The parser employs four modules to decide the actions to take:
\begin{enumerate}
\item a transition classifier $\theta(\cdot)$: a feed-forward neural network to decide the transition action to take, among $\Shift, \LArc, \RArc$ and $\Reduce$, based on the current configuration);

\item a concept identifier $\chi(\cdot)$: a dictionary extracted from the training dataset which decides the concept sub-graph (if any) related to a token to attach, once a $\Shift$ action takes place);

\item an edge labelling classifier $\lambda(\cdot)$: another feed-forward neural net for deciding the relation labels, e.g., ``:ARG0", ``:location", etc, between the top-two nodes of the stack after either a $\LArc$ or $\RArc$ action takes place;

\item a reentrancy classifier $\rho(\cdot)$: implemented again by a feed-forward neural net which decides whether to create a reentrant edge after a $\Reduce$ action).
\end{enumerate}

  \begin{wrapfigure}{R}{0.55\textwidth}
    \begin{minipage}{.55\textwidth}
      \begin{algorithm}[H]
        \caption{\textsc{AmrEager} Parsing}
        \label{alg:parsing}
          \begin{algorithmic}
            \STATE {\bfseries Input:} sentence $x$
            \STATE {\bfseries Return:} AMR graph $G$
            \STATE $state = \{\text{stack}\leftarrow[\text{root}],\text{buffer}\leftarrow[x],\text{edges}\leftarrow[]\}$
            \REPEAT
            \STATE action = $\theta(state)$
            \IF {$action ==\LArc |\RArc$}
            \STATE $label = \lambda(state)$
            \STATE $state = \textsc{APPLY}(action, label, state)$
            \ENDIF
            \IF {$action == \Shift$}
            \STATE $subgraph = \chi(state)$
            \STATE $state = \textsc{APPLY}(action, subgraph, state)$
            \ENDIF
            \IF {$action == \Reduce$}
            \STATE $reentrancy = \rho(state)$
            \STATE $state = \textsc{APPLY}(action, reentrancy, state)$
            \ENDIF
            \UNTIL{both \text{stack and buffer} are empty}
            \STATE {\bfseries return} build\_graph(state)
          \end{algorithmic}
      \end{algorithm}
    \end{minipage}
  \end{wrapfigure}
  
For the parser to decide what actions to apply to a configuration, feature templates are designed to be applied to the top nodes of the stack and/or the top tokens in the buffer.
These features can be the depth of the subgraph of nodes, the number of parent/child nodes, lexical embeddings for nodes/tokens, POS tags, name-entities for nodes/tokens or dependency relations among the nodes/tokens. The overall algorithm is shown in \cref{alg:parsing}.

For training these classifiers, an oracle similar to \citet{chen2014fast}'s is employed to compute transition sequences given AMR labeled sentences. Given any configuration of the parse, the oracle uses a fixed set of rules to determine the action to take (check \cite{damonte-17} for more detail).

\paragraph{Contextual Word Embeddings} Pre-trained BERT \cite{devlin2018bert} and RoBERTa \cite{liu2019roberta}\footnote{The models implementations are obtained from: https://huggingface.co/transformers/pretrained\_models.html} are used to extract word embeddings as part of the input feature for parsing. 
Both BERT and RoBERTa use a multi-layer bidirectional Transformer encoder architecture almost identical to \citet{vaswani2017attention}'s implementation, where each encoder layer includes two sub-layers; one multi-head self-attention layer followed by a fully connected feed-forward layer.
BERT is trained on a mixture of masked language modeling and next sentence prediction objectives; while RoBERTa is trained with dynamic masking during training, without the next sentence prediction objective, but with some other minor differences.
Although they are designed chiefly for the fine-tuning approach of pre-training general language representations, many have demonstrated their utility in the feature-extraction approach due to their ability to extract contextual information from the whole sentence, which is another motivation for using them with \textsc{AmrEager}.

\section{AMR parsing with contextual word embeddings}\label{sec:design}
We build on \textsc{AmrEager} to be able to incorporate contextual word features, and we dub the new system as \textsc{AmrBerger}.

These contextual word-features are prepared for training/testing in the pre-processing stage alone with the named-entities, dependency trees, POS tags for the whole dataset.
To generate the particular training/validation dataset for the neural-classifiers, the complete 768-dimensional (or 1024-dimensional in case of BERT$_\text{LARGE}$ and RoBERTa$_\text{LARGE}$) word embeddings are recorded along with the other input features, and the target label. This is in contrast to only having to record the word embedding index in a lookup table in the case of non-contextual embeddings.

\paragraph{Alignment method.} Due to both BERT and RoBERTa require splitting word-level tokens in sentences to combinations of sub-word units in their vocabulary--WordPiece \cite{wu2016google} and Byte-Pair Encoding (BPE) \cite{sennrich2015neural} in BERT and RoBERTa respectively. 
We have to find an alignment between these sub-word units and the word units used in our AMR aligner, dependency parser, etc.
The simple alignment algorithm we employ is: 
feed the BERT/RoBERTa tokenizer the tokens generated from the syntactic annotation pipeline one-by-one, and calculate, according to this output, the span of hidden vectors in the transformer output corresponding to each token. 
After than, the spans of hidden vectors corresponding to word-level units can be canonicalized by methods such as averaging or picking the head or last vectors, etc. Different methods are experimented in \cref{sec:experiments}.

In summary, \textsc{AmrBerger} differs from \textsc{AmrEager} mainly in that:
\begin{enumerate}
\item In preprocessing, an extra step is required to prepare the contextual token embeddings, for the tokens output from the CoreNLP tokenization pipeline \cite{manning2014stanford};

\item The training/validation dataset generation procedure for the neural classifiers are modified to record the contextual embeddings directly, instead of storing the word indices in the embedding lookup table, which does not contain any information about the context. 
As a drawback of this approach, each entry in the training/validation dataset is $(d-1)*n$ times larger than the original ones, with $d$ denotes the number of dimensions for each embedding, $n$ the number of tokens/nodes in each entry;

\item For the classifiers, the original table lookup module at the input layer is replaced with an identity module to pass on the pre-computed embeddings.
\end{enumerate}

We note that for $\amreager$ the lexical features for the neural classifiers' input not only consist of embeddings of 
\begin{inparaenum}
\item certain tokens on the buffer;
\item the tokens that certain nodes on the stack are aligned to,
\end{inparaenum} 
but also embedding of the concepts of the selected stack nodes. 
These concept embeddings (e.g., ``United States" might have a concept label ``country") are, as we will see in Section \ref{sec:experiments}, critical to the performance of the parsers. 
On one hand, we hypothesize this phenomenon is due to imperfect alignment between word tokens and AMR nodes, or abstraction artifact when generating the concept nodes for a sentence.
On the other, this illustrates an inherited limitation of using solely contextual embeddings; there is no straightforward way of obtaining the embeddings for these concept words. 
So one way to mitigate is to use the non-contextual embeddings for concepts, in conjunction with contextual embeddings for sentence words, which is experimented in \cref{sec:experiments}.

\section{Experiments}\label{sec:experiments}
$\amrberger$ with different transformer sizes, hidden-layer choices, and embeddings selection methods are experimented.
The best performing model is compared with the baseline $\amreager$ using the fine-grained Smatch evaluation.
We use the labelling classifier's performance as a heuristic in choosing the embedding configuration due to the following 4 reasons:
\begin{inparaenum}
\item It requires combinatorially more compute to find a global optimal hyperparameter setting suitable for all three neural nets, e.g., the action classifier alone has more than $1e6$ training samples thus takes a long time to converge;
\item Empirically, we found that under different settings, the performance of the three neural nets are highly positively correlated;
\item The labelling task is the hardest task of the three--having to classify among more than 100 potential labels, and it is where \textsc{AmrEager} performs the worst--thus an improvement under this task should bring a sizable gain in the overall SMATCH score.
\end{inparaenum}

For reference in the following experiments, $\amreager$ achieves an accuracy of 79.8\% on the labelling-classification development set.

\subsection{Transformer configuration}
Let $L$ denotes the number of Transformer encoder layers, $H$ the hidden layer dimension, $A$ the number of self-attention heads, $V$ the vocabulary cardinality. 
The following 5 pre-trained BERT and RoBERTa models are experimented in our system: 
BERT$_\text{BASE\_UNCASED}$ ($L=12, H=768, A=12, V=30,000$), BERT$_\text{BASE\_CASED}$ ($L=12, H=768, A=12, V=30,000$), BERT$_\text{LARGE\_CASED}$ ($L=24, H=1024, A=16, V=30,000$), RoBERTa$_\text{BASE}$ ($L=12, H=768, A=12, V=50,000$), and  RoBERTa$_\text{LARGE}$ ($L=24, H=1024, A=16, V=50,000$). 
At the time of design, the smaller BERT models were mainly designed to compare and contrast with OpenAI's GPT model \cite{radford2018improving} with a similar number of parameters, while the bigger ones are developed to assess the architecture's full potential.
The difference between ``CASED" and ``UNCASED" model is ``UNCASED" would treat tokens with capital letters indifferently with them with lower case letters (e.g. treating ``Apple", commonly representing the company's name, as ``apple", refering to the fruit). 
On the other hand, since their $V$ equals, ``UNCASED" would include more whole-word tokens and less out-of-vocabulary tokens.
With each pretrained model, the hyperparameters of each neural classifier are optimized through random search \cite{bergstra2012random}.
Across all transformer configurations, the penultimate layer of hidden activations is extracted for word embeddings.

\begin{table}[h]
    \caption{The labelling classifier accuracy under different configurations.}
    \vskip 0.15in
    \centering
    \begin{tabular}{|l|c|}
    \toprule
    Configuration & Accuracy \\ 
    \midrule
    BERT$_{\text{BASE\_UNCASED}}$  & 72.3\%\\
    BERT$_{\text{LARGE\_UNCASED}}$ & 74.6\%\\
    BERT$_{\text{LARGE\_CASED}}$ & 74.7\%\\
    RoBERTa$_{\text{BASE}}$ & 76.1\%\\
    RoBERTa$_{\text{LARGE}}$ & \textbf{77.5}\%\\
    \bottomrule
    \end{tabular}
    \vskip -0.1in
    \label{tab:size}
\end{table}

The performance of different contextual embeddings configurations is listed in \cref{tab:size} This agrees with their performance from previous evaluations \cite{liu2019roberta}; 
larger models perform better than their smaller counterparts, and RoBERTa outperforms BERT, confirming the previous finding that RoBERTa is a robustly optimized configuration of BERT.
In terms of the hyperparameters for the classifier, the best performance is obtained with a six-layer net each with 768-dim hidden layers, and trained using gradient descent with momentum. 

\subsection{Different activation layers}

As shown in \cite{devlin2018bert}, a different layer of activates would lead to noticeable performance difference in the Named Entity Recognition task. 
Two different activation-extraction methods are examined here with the RoBERTa$_\text{LARGE}$ model, the better performing model from the experiment above. 
The extraction methods are a) use the penultimate layer directly and b) extract the last four hidden layers and compute the arithmetic mean among the layers. 
Hyperparameter optimization is performed again in each of the settings.

\begin{table}[h]
    \caption{The labelling classifier accuracy using different activation-extraction method from RoBERTa$_\text{LARGE}$.}
    \vskip 0.15in
    \centering
    \begin{tabular}{|l|c|}
    \toprule
    Configuration & Accuracy \\ 
    \midrule
    Second-to-Last Hidden & \textbf{77.3}\%\\
    Average of Last Four Hidden & 77.0\%\\
    \bottomrule
    \end{tabular}
    \vskip -0.1in
    \label{tab:diff_layer}
\end{table}

The results in \cref{tab:diff_layer} show the score of the average method is a slightly lower than its alternative. 
This is the different from the finding from \cite{devlin2018bert} with BERT, which could be explained by the different nature of the tasks and could be informative to future design decisions.

\subsection{Different averaging method}

As introduced in \cref{sec:design}, after obtaining the spans of sub-word tokens corresponding to a word-level token, different methods exist to compute the final feature vectors representing the word-level tokens. 
In our experiments, the headword token's embeddings and the average of the spanned sub-word embeddings from RoBERTa$_\text{LARGE}$'s penultimate layer are evaluated.

\begin{table}[h]
    \caption{The labelling classifier accuracy using different method of resolving multiple sub-word level units corresponding to one word level unit from RoBERTa$_\text{LARGE}$.}
    \vskip 0.15in
    \centering
    \begin{tabular}{|l|c|}
    \toprule
    Configuration & Accuracy \\ 
    \midrule
    Head Token of the span & 76.9\%\\
    Average of Tokens in the span & \textbf{77.3}\%\\
    \bottomrule
    \end{tabular}
\vskip -0.1in
    \label{tab:average-method}
\end{table}

Results in \cref{tab:average-method} shows that averaging the spanning tokens' performs slightly better than only taking the head token, which is also what \citet{devlin2018bert} employed when extracting BERT embeddings for the named-entities-recognition task. 
The averaging method seems to contain more discriminative information for the classifier than just the head token method, though in theory, each token should have attended to all the other context tokens.

\subsection{Additional Concept Features}
The results above demonstrate, even in the best condition, the labelling classifier using purely contextual word vectors obtains $77\%$ accuracy on the dev set, while the one using non-contextual word vectors achieves $3\%$ higher than that, as shown in \cref{tab:performance}. 
We attribute this poor performance to the absence of concept embeddings in \textsc{AmrBerger}. 
This situation is effectively rescued by leveraging additional GloVe embeddings for node concepts on top of the contextual embeddings for words, as shown in the third column of \cref{tab:performance}. 
We observe that with the help of these concept embeddings, \textsc{AmrBerger} significantly improves in labelling classification, outperforming \textsc{AmrEager}. 

\textbf{The Other Classification Tasks and Fine-grained Smatch comparison}
\cref{tab:performance} tabulates the performance of the other classifiers and the fine-grained Smatch score of the best performing configuration of \textsc{AmrEager}, \textsc{AmrBerger} and \textsc{AmrBerger} plus concept features.

We observe that albeit \textsc{AmrBerger}'s poor accuracy in labelling and reentrancy classification in comparison to \textsc{AmrEager}, it outperforms \textsc{AmrEager} by $4\%$ in transition classification, the most frequently used component of the system. 
As a result, they achieve the same overall Smatch statistics. 
The mixed fine-grained Smatch performance agree with the strengths and weaknesses of each system's classifiers. 
With additional concept features, \textsc{AmrBerger}+Concept outperforms \textsc{AmrEager} in all metrics except in reentrancy, though only by a small margin. In the end, its Smatch score is 2 points higher than \textsc{AmrEager}.

\begin{table*}[!h]
\caption{Performance of the models on the pre-split development set of the LDC2020T02 dataset. $\alpha$(Transition), $\alpha$(labelling), $\alpha$(reentrancy), stands for the classifiers' accuracy on the development set. The others are the SMATCH and its decomposition scores.} 
\label{tab:performance}
\vskip 0.15in
\begin{center}

\begin{tabular}{|l|c|c|c|}
\toprule
Metric               & \textsc{AmrEager} & \textsc{AmrBerger} & \textsc{AmrBerger}+Concept\\
\midrule
$\alpha$(Transition) & 84\%            & 88\%                 &  88\%\\
$\alpha$(labelling)   & 80\%            & 77\%                 &  82\%\\
$\alpha$(Reentrancy) & 98\%            & 96\%                 &  97\%\\
\midrule
Smatch              & 63            &   63                    &  65\\
Unlabeled           & 67            &   68                    &  68\\
No WSD              & 63            &   64                    &  64\\
Reentrancy          & 45            &   44                    &  46\\
Concepts            & 81            &   81                    &  81\\
Named Ent.          & 78            &   77                    &  80\\
Wikification        & 61            &   61                    &  61\\
Negations           & 46            &   46                    &  46\\
SRL                 & 61            &   61                    &  63\\
\bottomrule
\end{tabular}

\end{center}
\vskip -0.1in
\end{table*}

\subsection{Ablation Study}\label{sec:ablation}
As analyzed in \cite{liu-etal-2019-linguistic, tenney-etal-2019-bert, tenney2018what}, contextual word embeddings, especially BERT, can capture strong syntactical information about words. 
We therefore experiment with a variation of \textsc{AmrEager} and \textsc{AmrBerger} that does not use explicit syntactic dependency features, to probe how the use of contextual embeddings would help in this situation.

\begin{table}[]
    \caption{Different system's labelling classification accuracy on the dev set, all without dependency features.}
    \vskip 0.15in
    \centering
    \begin{tabular}{|l|c|}
    \toprule
    Configuration & Accuracy \\ 
    \midrule
    \textsc{AmrEager}-dep & 77.4\%\\
    \textsc{AmrBerger}-dep & 75.9\%\\
    \textsc{AmrBerger}-dep+concept & 81.1\%\\
    \bottomrule
    \end{tabular}
\vskip -0.1in
    \label{tab:ablations}
\end{table}

We observe from \cref{tab:ablations} that even without dependency features, \textsc{AmrEager} has higher accuracy than when \textsc{AmrBerger} has the explicit dependency signals. 
On the other hand, the accuracy of the classifiers using contextual word embeddings suffer less from the removal of the dependency features.

\section{Conclusions}\label{sec:conclusions}
In this study, we found contextual embeddings alone are not able to provide much performance improvements in our transition based AMR parser, as demonstrated by the classifiers' accuracy and the final Smatch scores.
This is not entirely consistent with the kind of performance gain shown in much of the other works in the space, indicating potential limitations of the current approach, and encourages us to understand deeper about the nature of these pretrained models in future works.
A hybrid approach with contextual embedding for words and non-contextual embeddings for concepts achieves the best performance, highlighting the importance of employing the concept features in AMR parsing.
Nonetheless, as demonstrated by the ablation study, contextual embeddings do show evidence of capturing richer information about the sentence, namely syntactical knowledge of words.

\section{Acknowledgment}
I thank my advisor Shay B. Cohen for his support and insightful discussion throughout this project.

\bibliographystyle{abbrvnat}
\bibliography{main}

\end{document}